\documentclass[journal]{IEEEtran}
\IEEEoverridecommandlockouts
\usepackage{cite}
\usepackage{amsmath,amssymb,amsfonts}
\usepackage{algorithmic}
\usepackage{graphicx}
\usepackage{textcomp}
\usepackage{xcolor}
\usepackage{array}
\usepackage{dblfloatfix}
\usepackage{hyperref}
\usepackage{float}
\hypersetup{colorlinks,linkcolor={blue},citecolor={blue},urlcolor={red}}  

\def\BibTeX{{\rm B\kern-.05em{\sc i\kern-.025em b}\kern-.08em
    T\kern-.1667em\lower.7ex\hbox{E}\kern-.125emX}}
    
\begin{document}

\makeatletter
\newcommand{\linebreakand}{%
\end{@IEEEauthorhalign}
\hfill\mbox{}\par
\mbox{}\hfill\begin{@IEEEauthorhalign}
}
\makeatother

\def\titulo{A New State-of-the-Art \textit{Transformers}-Based Load Forecaster on the Smart Grid Domain}

\title{\titulo}

\author{André~Luiz~Farias~Novaes,~\IEEEmembership{Student Member,~IEEE,}
        Rui~Alexandre~de~Matos~Araújo,~\IEEEmembership{Senior Member,~IEEE,}
        José~Figueiredo,~and~Lucas Aguiar Pavanelli,~\IEEEmembership{Student Member,~IEEE}

\thanks{Manuscript received on \ldots. Accepted for publication \ldots.

A.L.F.~Novaes, and R.A.M.~Araújo are with the Institute for Systems and Robotics (ISR-UC), and Department of Electrical and Computer Engineering (DEEC-UC), University of Coimbra, Pólo II, PT-3030-290 Coimbra, Portugal. (e-mails: andre.novaes@isr.uc.pt, rui@isr.uc.pt).

J.~Figueredo is with the Critical Software S.A., Pq.\ Ind.\ de Taveiro, Lt 49, PT-3045-504 Coimbra, Portugal. (e-mail: jose.figueiredo@criticalsoftware.com).

L.A.~Pavanelli, and A.L.F.~Novaes are with Department of Computer Science, PUC-Rio, Rio de Janeiro, Brazil (e-mail: lpavanelli@inf.puc-rio.br).

} 
}

\markboth{IEEE ..... ........... Letters,~Vol.~XX, No.~Y, Month~Year}%
{Novaes \MakeLowercase{\textit{et al.}}: \titulo}


\maketitle

\begin{abstract}
Meter-level load forecasting is crucial for efficient energy management and power system planning for Smart Grids (SGs), in tasks associated with regulation, dispatching, scheduling, and unit commitment of power grids. Although a variety of algorithms have been proposed and applied on the field, more accurate and robust models are still required:\ the overall utility cost of operations in SGs increases 10 million currency units if the load forecasting error increases 1\%, and the mean absolute percentage error (MAPE) in forecasting is still much higher than 1\%. Transformers have become the new state-of-the-art in a variety of tasks, including the ones in computer vision, natural language processing and time series forecasting, surpassing  alternative  neural  models  such  as convolutional and recurrent neural networks. In this letter, we present a new state-of-the-art Transformer-based algorithm for the meter-level load forecasting task, which has surpassed the former state-of-the-art, LSTM, and the traditional benchmark, vanilla RNN, in all experiments by a margin of at least 13\% in MAPE.
\end{abstract}

\begin{IEEEkeywords}
Transformers, deep learning, long-short term memory, meter-level load forecasting, short-term load forecasting.
\end{IEEEkeywords}

\section{Introduction}


\IEEEPARstart{E}{lectrical} load forecasting plays a crucial role on the Smart Grid (SG) domain, particularly on efficient energy management and power system planning~\cite{zheng2013smart,avancini2019energy,depuru2011smart}. Consequently, high-accuracy forecasts are required in multiple time horizons for tasks associated with regulation, dispatching, scheduling, and unit commitment of power grids~\cite{raza2015review,hafeez2020fast}.

The meter-level load forecasting has become an active research topic, with a variety of algorithms applied on it:\ \cite{hafeez2020electric, amjady2010short, ahmad2016accurate, hafeez2020fast, zeng2017switching, sideratos2020novel, zhang2017short, kong2017short, zhang2016composite, al2013multivariate, zhang2013short, ghofrani2015hybrid, shi2017deep, kong2017short1, kaur2019smart, nugaliyadde2019predicting}, and others. \textit{However, more accurate and robust models are still required in meter-level load forecasting} ~\cite{hafeez2020electric}, mainly due the following reasons:\ external factors influencing forecasts, e.g.\ climate, calendar events, occupancy patterns, social conventions; the stochastic and non-linear behavior of consumers; the sensitive decision making process and operation of the SG. There is a 10 million currency units increase in the overall utility cost if the load forecasting error increases in 1\%.

The \textit{Transformer} deep learning architecture \cite{vaswani2017attention} has become the new state-of-the-art forecaster in a variety of tasks \cite{katharopoulos2020transformers, tay2020efficient, dosovitskiy2020image, pfeiffer2020adapterhub, zaheer2020big, wolf2020transformers}, including on natural language processing, surpassing alternative neural models such as CNNs and RNNs. The Transformers' architecture scales with training data and model size, facilitates efficient parallel training, and captures long-range sequence features~\cite{wolf2019huggingface}. Unlike sequence-aligned models, the Transformer does not process data in an ordered sequence manner: it processes entire sequence of data and uses \textit{self-attention} mechanisms to learn dependencies in the sequence. Therefore, Transformer-based models have the potential to model complex dynamics of time series data that are challenging for sequence models~\cite{wu2020deep}.

\textit{The main contribution of this letter is to present a new state-of-the-art Transformer-based algorithm for the meter-level load forecasting task}. Although it has been successfully deployed in many real-world applications and recently developed for time series forecasting (TSF)~\cite{wu2020deep, li2019enhancing, oh2018learning, lim2019temporal}, \textit{Transformers has not yet been applied in the electricity load forecasting community}. In this letter, we fill this gap, specifically applying it on \textit{The Smart Meters in London Dataset}, one of the largest public-domain datasets on SGs, the largest one hosted at Kaggle\footnote{https://www.kaggle.com/jeanmidev/smart-meters-in-london}.

\section{The Method}
\label{s.method}

Our Transformer-based forecaster follows the one developed in~\cite{wu2020deep}, which is in turn based on the original architecture of~\cite{vaswani2017attention}, minutely described in Figure \ref{fig1}.

\begin{figure}[t]
    \centering
    \includegraphics[width=0.7\columnwidth]{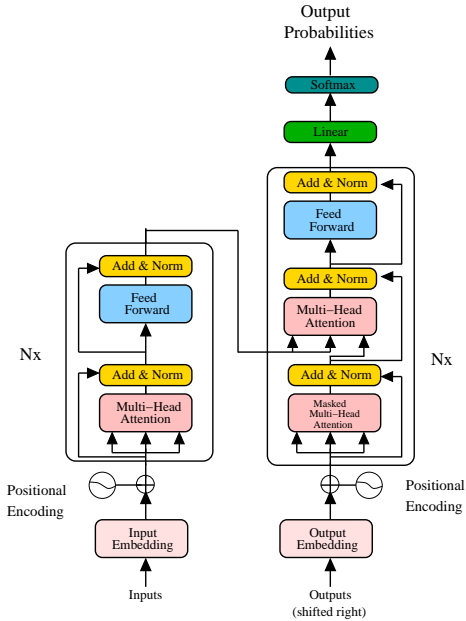}
    \caption{The Transformer Model Architecture \cite{vaswani2017attention}.}
    \label{fig1}
\end{figure}\relax

The Transformer is a encoder-decoder architecture. On a high level, the encoder maps an input sequence, e.g.\ a time series sequence, into an abstract continuous representation that holds all information learned from the input. The decoder takes that continuous representation and, step-by-step, generates a single output while also feeding the previous output.


The encoder, the left block of Figure \ref{fig1}, is composed of a stack of $N = 6$ identical layers, with two sub-layers each:\ the first layer is a multi-head \textit{self-attention} mechanism; and the second layer is a fully connected feed-forward network. Residual connections \cite{he2016deep} and normalization operations \cite{ba2016layer} are applied on each sub-layer.


The decoder, the right block of Figure \ref{fig1}, is similar to the encoder. The main difference, not the only one, is an extra sub-layer on the stack, \textit{Masked Multi-Head Attention}.

The Transformer architecture mainly differentiates itself from other deep networks created before by \textit{self-attention}, a mechanism that balances long-range dependencies modeling and computational efficiency~\cite{zhang2019self}. The full self-attention module calculates response at a position as a weighted sum of the features at all positions. Unlike the RNN-based methods, Transformers allow access to any part of the time series history regardless of distance, potentially grasping recurring patterns with long-term dependencies~\cite{li2019enhancing}.

\section{Experiments and Results}
\label{s.expresul}

\textit{The Smart Meters in London Dataset} is composed of 5567 London Households energy consumption readings between November/2011 and February/2014, with a 30 minutes frequency. For this letter, we have selected for forecasting the same eight houses\footnote{The selected houses are: MAC000002, MAC000033, MAC000092, MAC000156, MAC000246, MAC000450, MAC001074, and MAC003223.} as in \cite{nugaliyadde2019predicting}, each of them with its' own number of samples.

The dataset is split into training and test sets, with a 80\%-20\% ratio. The inputs to the prediction model are the measurements of the consumption on the last $n$ time intervals (TI), from the current TI $k$ to the past, i.e.\ the inputs are the consumption measurements on intervals $k-n+1, \ldots, k$; and the output of the model is the consumption forecast for the next TI, $k+1$, in kWh.

Experiments were conducted on the Amazon Web Services (AWS) ml.p2.xlarge instance, 4 vCPUs, 1xK80 GPU, 61 GB of memory, and 12 GB of GPU memory, totaling approximately 310 hours of GPU-time. The proposed transformer-based algorithm is compared against the former state-of-the-art in residential load forecasting \cite{kong2017short1}, LSTM, and the vanilla RNN, traditionally used as a benchmark in the TSF domain. Since all the models rely on random initialization of network weights, five experiments were performed for each of the three models, for all eight houses, training using four distinct values for the number $n$ of TIs, specifically for $n=2,3,6,12$. Therefore, the total number or runs is 480.

\begin{table}[t]
\caption{Load Forecasting Summary.}
\label{t.resul}
\vspace{-1.0em}
\begin{center}
\begin{tabular}{| m{2.2cm} | m{1.5cm} | m{2.8cm} |}
\hline
\textbf{Algorithm} & \textbf{MAPE Avg.} & \textbf{Total Train. Time [sec]} \\
\hline
Transformer-2TI & 64.87\% & 11.53668 \\
\hline
LSTM-2TI &  82.62\% & 3.04872 \\ 
\hline
RNN-2TI & 78.50\% & 2.08464 \\
\hline
Transformer-3TI & 62.75\% & 10.68004 \\
\hline
LSTM-3TI &  75.38\% & 3.09136 \\ 
\hline
RNN-3TI & 79.62\% & 2.11112 \\
\hline
Transformer-6TI & 62.87\% & 7.84844 \\
\hline
LSTM-6TI & 80.50\% & 3.08364 \\ 
\hline
RNN-6TI & 81.25\% & 2.10912 \\
\hline
Transformer-12TI & 63.50\% & 8.42956 \\
\hline
LSTM-12TI & 81.88\% & 3.09936 \\ 
\hline
RNN-12T & 76.25\% & 2.11384 \\
\hline
\end{tabular}\\
Note: all the time values have been multiplied by $10^{-4}$ in the column \textbf{Total Train. Time [sec]} .
\label{tab1}
\end{center}
\end{table}\relax

Overall results are shown in Table \ref{t.resul}, where \textbf{MAPE Avg.} is the average of the mean absolute percentage error (MAPE) over all the 40 experiments performed in the test sets for the configuration (type of model, and number of TIs) detailed in the corresponding line of Table \ref{t.resul}, and \textbf{Total Train. Time [sec]} is the total training time, in seconds, for the same 40 experiments for each line. The results show that the Transformer-based algorithm outperforms LSTM and RNN in all experiments by a margin of at least 13\% in MAPE.

Forecasting electric load for single residential users is much more changeable than for aggregated residential loads in a large scale or even industrial consumers, mainly due the high volatility and uncertainty involved, generated by distinct daily routines residents' lifestyle; industrial electricity consumption patterns, for example, are much more regular than residential ones \cite{kong2017short}. Results achieved at Table \ref{tab1} are comparable in magnitude with results of leader works in the \textit{smart grid forecast} domain, as in \cite{kong2017short}.

Figure \ref{fig3} 
\begin{figure}[t]
\centerline{\includegraphics[width=0.95\columnwidth]{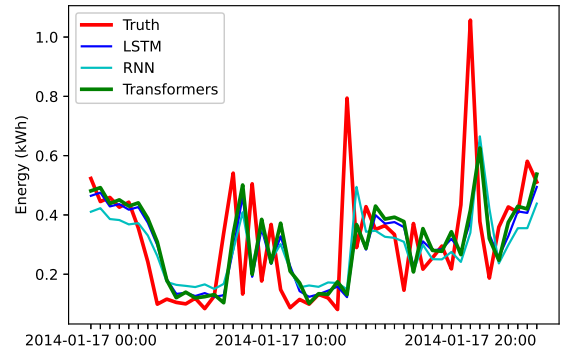}}
\caption{Forecasting energy for house MAC000002.}
\label{fig3}
\end{figure}\relax
shows how each of the methods perform on house MAC000002 in a specific day. Qualitatively, it is possible to observe that Transformers better responds to volatile movements in the load series, identified by ``Truth''.

\section{Conclusion}

The relevance of a new state-of-the-art Transformer model in the meter-level load prediction is remarkable. The results indicate that the Transformer learning model achieves superior performance when compared to LSTM and RNN models for meter-level load prediction. Created in 2017, Transformers are allowing new possibilities for TSF in distinct areas, with pure and hybrid models. Presented to the load forecasting community in 2018 as a new state-of-the-art \cite{kong2017short1}, LSTM networks are still valuable forecasters, with at least one advantage over Transformers:\ the smaller amount of training time. Next research directions will explore ways on developing less costly Transformers models in terms of training time.

\section*{Acknowledgment}

This work was supported by Project CONNECTA-X/2017/33354 co-financed by PT2020, in the framework of the COMPETE 2020 Programme, and by the European Union through the European Regional Development Fund (ERDF).
\vspace{-0.7em}
\begin{center}
\includegraphics[width=5.0em]{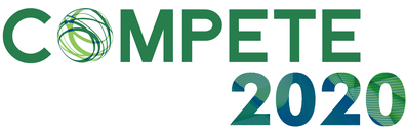}
\hspace{0.3em}
\includegraphics[width=5.0em]{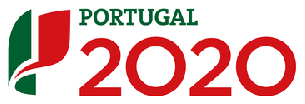}
\hspace{0.3em}
\includegraphics[width=5.4em]{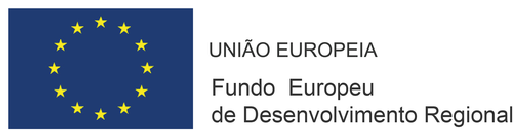}
\end{center}
\vspace{-0.3em}

\vspace{12pt}

\bibliographystyle{IEEEtran}
\bibliography{\jobname}



\end{document}